\documentclass[10pt,twocolumn,letterpaper]{article}

\usepackage{cvpr}

\usepackage{cite}
\usepackage[latin9]{inputenc}
\usepackage{amsthm}
\usepackage{amsmath}
\usepackage{amssymb}
\usepackage{enumerate}
\usepackage{centernot}
\usepackage{xspace}
\usepackage{enumitem}
\usepackage{environ}
\usepackage{times}
\usepackage{float}
\usepackage{algorithmicx}
\usepackage{algpseudocode}
\usepackage{bm}
\usepackage{lscape}
\usepackage{pdflscape}
\usepackage{wrapfig}
\usepackage{rotating}
\usepackage{epstopdf}
\usepackage{setspace}
\usepackage{breqn}

\usepackage{epsfig}
\usepackage{xcolor}

\usepackage{algorithm}

\usepackage[hyphens]{url}

\usepackage{graphicx,subcaption}

\usepackage{multirow}

\usepackage{array}
\newcolumntype{P}[1]{>{\centering\arraybackslash}p{#1}}
\newcolumntype{M}[1]{>{\centering\arraybackslash}m{#1}}

\newcommand{\lyxmathsym}[1]{\ifmmode\begingroup\def\b@ld{bold}
	\text{\ifx\math@version\b@ld\bfseries\fi#1}\endgroup\else#1\fi}

\usepackage[breaklinks=true,bookmarks=false]{hyperref}

\cvprfinalcopy 

\def\cvprPaperID{****} 

\begin{document}
	
\title{Deceiving Google's Cloud Video Intelligence API Built for Summarizing Videos}


\author{Hossein Hosseini \qquad Baicen Xiao \qquad Radha Poovendran\\
	Network Security Lab (NSL) \\
	Department of Electrical Engineering, University of Washington, Seattle, WA \\
	{Email: \{hosseinh, bcxiao, rp3\}@uw.edu}\\
}

\maketitle

\begin{abstract}

Despite the rapid progress of the techniques for image classification, video annotation has remained a challenging task. 
Automated video annotation would be a breakthrough technology, enabling users to search within the videos. 
Recently, Google introduced the Cloud Video Intelligence API for video analysis. 
As per the website, the system can be used to ``separate signal from noise, by retrieving relevant information at the video, shot or per frame'' level.
A demonstration website has been also launched, which allows anyone to select a video for annotation. The API then detects the video labels (objects within the video) as well as shot labels (description of the video events over time). 

In this paper, we examine the usability of the Google's Cloud Video Intelligence API in adversarial environments. 
In particular, we investigate whether an adversary can subtly manipulate a video in such a way that the API will return {\it only} the adversary-desired labels. 
For this, we select an image, which is different from the video content, and insert it, periodically and at a very low rate, into the video. We found that if we insert one image every two seconds, the API is deceived into annotating the video as if it only contained the inserted image. Note that the modification to the video is hardly noticeable as, for instance, for a typical frame rate of $25$, we insert only one image per $50$ video frames. We also found that, by inserting one image per second, all the shot labels returned by the API are related to the inserted image. 
We perform the experiments on the sample videos provided by the API demonstration website and show that our attack is successful with different videos and images.

\end{abstract}

\section{Introduction}

\begin{figure}[t]
	\centering
	\includegraphics[width=0.95\linewidth]{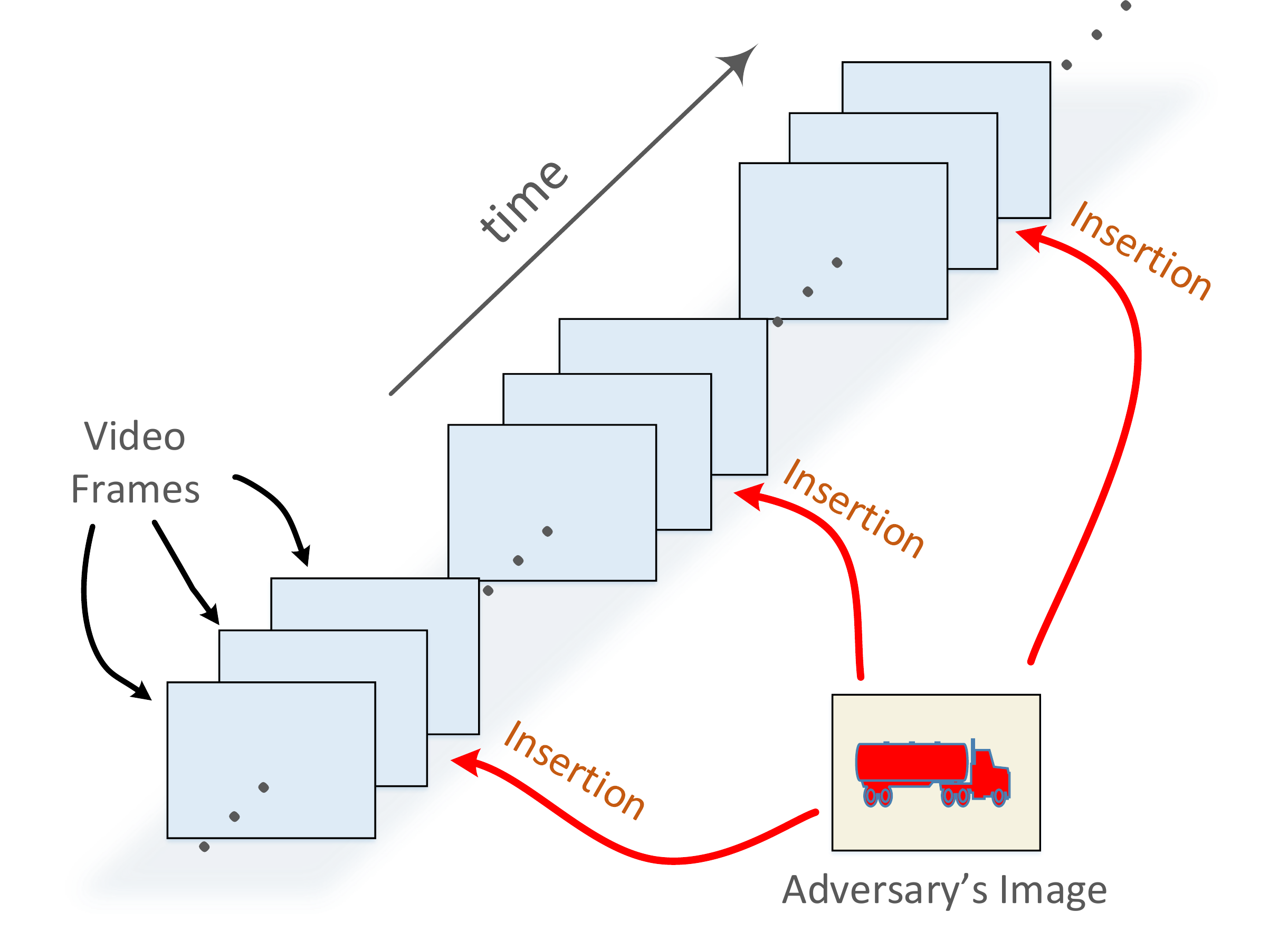}
	\caption{Illustration of the image insertion attack on Google's Cloud Video Intelligence API. The adversary chooses an image and inserts it, periodically and at a very low rate, into the video. Our experimental results show that, by inserting the image once every second, we can deceive the API to output only the labels of the inserted image for both the video and shot labels.}\vspace{-0.2cm}
	\label{fig:attack}
\end{figure}

In recent years, machine learning techniques have been extensively deployed for computer vision tasks, particularly recognizing objects in images~\cite{krizhevsky2012imagenet,simonyan2014very,girshick2014rich,he2016deep}. 
However, using machine learning for annotating videos has remained a challenging task, due to the temporal aspect of video data and the difficulty of collecting sufficient well-tagged training samples~\cite{vondrick2010efficiently,zha2012interactive,wang2012assistive}. 
Due to the growth of video data on the Internet, automatic video annotation has gained a lot of attention from the research community~\cite{wang2009unified,wang2009beyond,zhang2012generic}, as well as companies such as Facebook~\cite{facebook} and Twitter~\cite{twitter}. 
Automatic video annotation can enable searching the videos for a specific event, which is helpful in applications such as video surveillance or returning the search results on the web. It can be also used for prescanning user videos, for example in YouTube and Facebook, where  distribution of certain types of illegal contents is not permitted.

Recently, Google introduced the Cloud Video Intelligence API for video analysis~\cite{google}. 
A demonstration website has been launched which allows anyone to select a video stored in Google Cloud Storage for annotation~\cite{cloud}. The API then quickly identifies the {\it video labels} which are the key objects within the video. It also detects the scene changes and provides {\it shot labels} as the detailed description of the video events over time. Similar to other Google's machine learning APIs, the Cloud Video Intelligence API is made available to developers to build applications that can automatically search within the videos~\cite{google}. Hence, the API has the potential to simplify the video understanding and enable searching in videos just as text documents.

Machine learning systems are typically designed and developed with the implicit assumption that they will be deployed in benign settings. However, many works have pointed out their vulnerability in adversarial environments~\cite{barreno2006can,huang2011adversarial,papernot2016limitations,amodei2016concrete}. Security evaluation of machine learning systems is an emerging field of study. 
In~\cite{carlini2016hidden}, Carlini et al. showed that voice interfaces can be attacked with hidden voice commands that are unintelligible to humans, but are interpreted as commands by devices.
In~\cite{sharif2016accessorize}, Sharif et al. proposed techniques for physically realizable image modification to attack face-recognition systems. 
Recently, Hosseini et al. showed that the Google's Perspective API for detecting toxic comments can be defeated by subtly modifying the input text~\cite{hosseini2017deceiving}.

In this paper, we examine the usability of the Google's Cloud Video Intelligence API in adversarial environments. In particular, we investigate whether an adversary can deceive the API into returning {\it only} the adversary-desired labels, by slightly manipulating the input video. 
Such vulnerability will seriously undermine the performance of the video annotation system in real-world applications. For example, a search engine may wrongly suggest manipulated videos to users, or a video filtering system can be bypassed by slightly modifying a video which has illegal contents.

For manipulating the videos, we select an image, different from the video content, and insert it, periodically and at a very low rate, into the video. Our experimental results show that by inserting the image once every two seconds, the API is deceived into returning only the video labels which are related the inserted image.
Note that the modification to the video is hardly noticeable as, for instance, for a typical frame rate of $25$, we insert only one image per $50$ video frames. We also found that by inserting one image per second, all the shot labels returned by the API are related to the inserted image. 
We perform the experiments on the sample videos provided by the API demonstration website and with different images. 
Figure~\ref{fig:attack} illustrates the image insertion attack on the Google's Cloud Video Intelligence API.

\section{Google's Cloud Video Intelligence API}

%
%

The Google's Cloud Video Intelligence API is designed for video understanding and analysis. It enables the developers to easily search and discover the video content by providing information about entities (nouns or verbs) in the video and when they occur within the video. It was noted in~\cite{cloud} that the system can be used to ``separates signal from noise, by retrieving relevant information at the video, shot or per frame'' level. 
The API uses deep-learning models, built using frameworks such as TensorFlow and applied on large-scale media platforms such as YouTube~\cite{google}. 



The system is said to be helpful for large media companies to better understand the video data, and for media organizations and consumer technology companies, who want to build their media catalogs or find easy ways to manage crowd-sourced content~\cite{google}.
The underlying technology can be also used to improve the video recommendations, as it enables the search engines to consider the video content, beyond the metadata like descriptions and comments, for searches.

\begin{figure}[t]
	\centering
	\begin{subfigure}{.22\textwidth}
		\centering
		\includegraphics[width=1\linewidth]{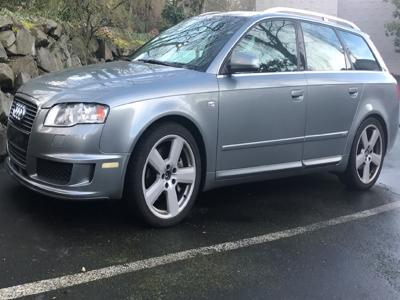}
		\caption{}
	\end{subfigure}
	\begin{subfigure}{.22\textwidth}
		\centering
		\includegraphics[width=1\linewidth]{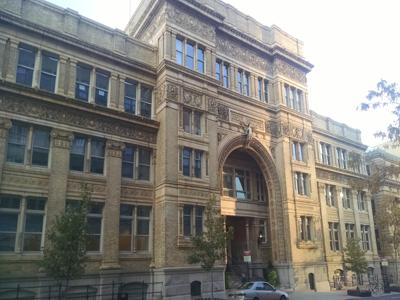}
		\caption{}
	\end{subfigure}\\\vspace{0.1cm}
	\begin{subfigure}{.22\textwidth}
		\centering
		\includegraphics[width=1\linewidth]{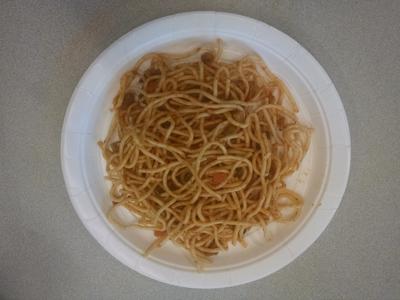}
		\caption{}
	\end{subfigure}
	\begin{subfigure}{.22\textwidth}
		\centering
		\includegraphics[width=1\linewidth]{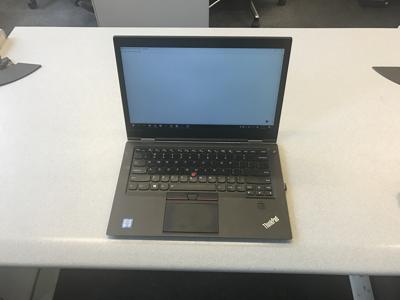}
		\caption{}
	\end{subfigure}
	\caption{The four images (a) a car, (b) a building, (c) a food plate, and (d) a laptop, that were used in experiments for inserting within the sample videos.}\vspace{-0cm}
	\label{fig:images}
\end{figure}


\bgroup
\def\arraystretch{1.22}
\begin{table*}[t]
	\centering
	\caption{Demonstration of the Image Insertion Attack on the Google's Cloud Video Intelligence API. We performed the experiments with three sample videos provided by the API website~\cite{cloud}. The images are inserted once every two seconds within the video, which is equal to inserting one image per $50$ video frames for a typical frame rate of $25$. For each of the input videos and inserted images, the table shows the video label with the highest confidence returned by the API.}
	\begin{tabular}{ |M{3.25cm}|M{3.5cm}|M{5.5cm}| } 
		\hline
		{{\bf Video Name}} & { {\bf Inserted Image}} & {\small {\bf Video Label Returned by API (Confidence Score)}} \\
		\hline
		\hline
		\multirow{4}{*}{{\bf ``Animals.mp4''}}
		& ``Car'' & Audi ($98\%$) \\ \cline{2-3} & ``Building'' & Building ($89\%$) \\ \cline{2-3} & ``Food Plate'' & Pasta ($99\%$) \\ \cline{2-3} & ``Laptop'' & Laptop ($91\%$)\\
		\hline
		\hline
		\multirow{4}{*}{{{\bf ``GoogleFiber.mp4''}}} & ``Car'' & Audi ($98\%$) \\ \cline{2-3} & ``Building'' & Classical architecture ($95\%$) \\ \cline{2-3} & ``Food Plate'' & Noodle ($99\%$) \\ \cline{2-3} & ``Laptop'' & Laptop ($91\%$)\\
		\hline
		\hline
		\multirow{4}{*}{{{\bf ``JaneGoodall.mp4''}}} & ``Car'' & Audi ($98\%$) \\ \cline{2-3} & ``Building'' & Classical architecture ($95\%$) \\ \cline{2-3} & ``Food Plate'' & Pasta ($99\%$) \\ \cline{2-3} & ``Laptop'' & Laptop ($91\%$)\\
		\hline
	\end{tabular}\label{table}
\end{table*}
\egroup


\section{The Image Insertion Attack}\label{sec:attack}

In this section, we describe the image insertion attack for deceiving the Google's Cloud Video Intelligence API. 
The goal of the attack is to modify a given video in such a way that a human observer would perceive its original content, but the API returns only the adversary-desired annotations. 
We performed the experiments with three sample videos ``Animals.mp4'', ``GoogleFiber.mp4'' and ``JaneGoodall.mp4'', which are provided by the demonstration website of the Google's Cloud Video Intelligence API~\cite{cloud}. 
The API provides video labels (objects in the entire video), shot changes (scene changes within the video) and shot labels (description of the video events over time).

The attack procedure is as follows.
We first tested the API with sample videos and verified that the API did indeed accurately detect both the video and shot labels. For example, for the ``Animals.mp4'' video, the API returns the video labels ``Animal,'' ``Wildlife,'' ``Zoo,'' ``Terrestrial animal,'' ``Nature,'' ``Tourism,'' and ``Tourist destination,'' which are consistent with the video content.

We then downloaded the sample videos and modify them. For manipulating the videos, we select an image, different from the video content, and insert it, periodically and at a very low rate, into the videos. Figure~\ref{fig:images} shows the four images that were used for image insertion attack, namely, a car, a building, a food plate and a laptop. The schematic of the image insertion attack is illustrated in Figure~\ref{fig:attack}. 
At the end, we stored the manipulated videos on the Google cloud storage and used them as inputs to the API.
\footnote{The experiments are performed on the interface of the Cloud Video Intelligence API's website on Mar. 24, 2017.} 

Our experimental results show that if we insert an image periodically once every two seconds and in appropriate places, the API completely fails to correctly understand the video content and annotates it as if the video was only about the inserted image. Note that the image insertion rate is very low. That is, for a typical frame rate of $25$, we insert only one image per $50$ video frames, resulting in an image insertion rate of $0.02$. Therefore, the modification to the video is hardly noticeable. 
Moreover, we tested the API with videos with different frame rates and verified that the attack is successful, regardless of the choice of the frame rate.

\begin{figure*}[h]
	\centering
	\begin{subfigure}{.8\textwidth}
		\centering
		\includegraphics[width=1\linewidth]{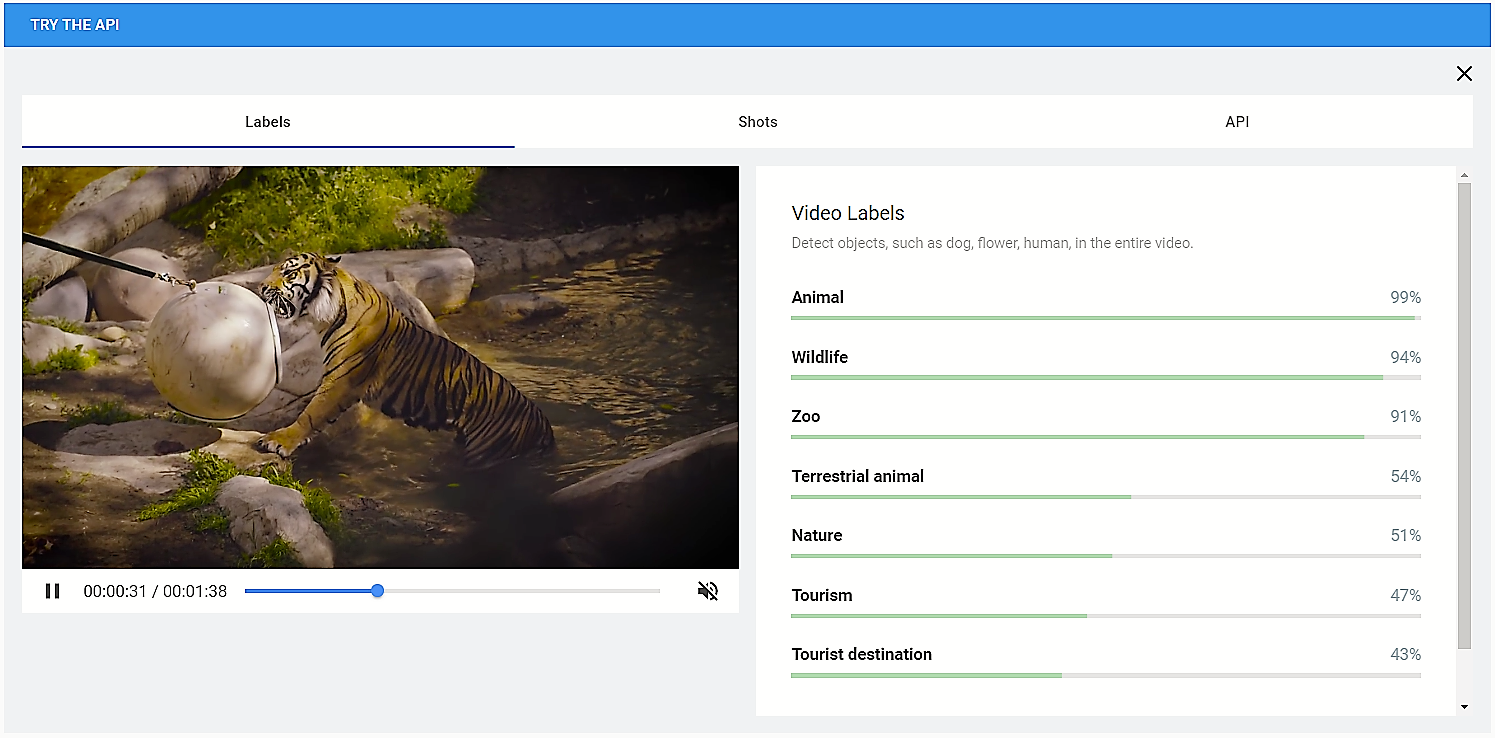}
		\caption{Video labels generated by API for the original video.}
	\end{subfigure}\\\vspace{0.5cm}
	\begin{subfigure}{.4\textwidth}
		\centering
		\includegraphics[width=1\linewidth]{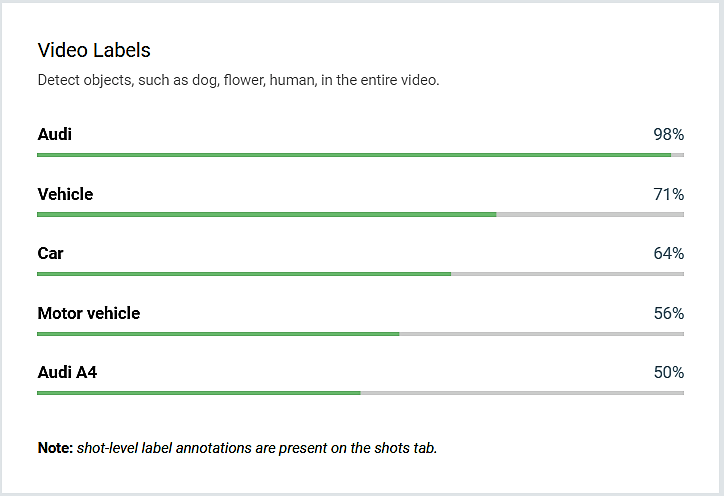}
		\caption{Video labels of the manipulated video, where an image of a car is inserted once every two seconds.}
	\end{subfigure}\hspace{1cm}
	\begin{subfigure}{.4\textwidth}
		\centering
		\includegraphics[width=1\linewidth]{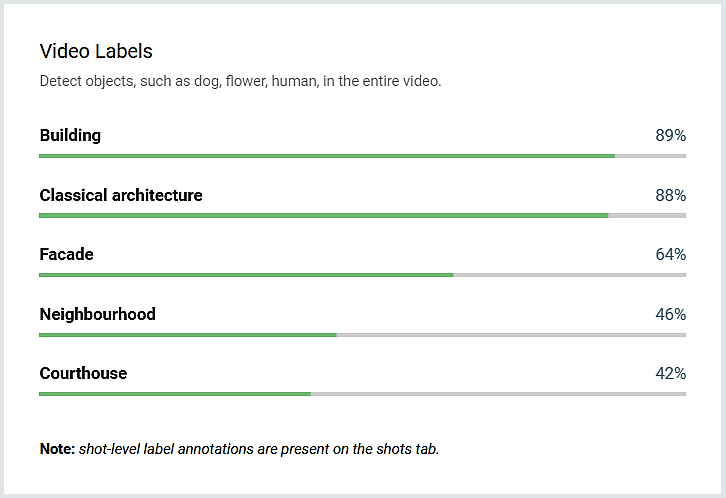}\vspace{0cm}
		\caption{Video labels of the manipulated video, where an image of a building is inserted once every two seconds.}
	\end{subfigure}\\\vspace{0.5cm}
	\begin{subfigure}{.4\textwidth}
		\centering
		\includegraphics[width=1\linewidth]{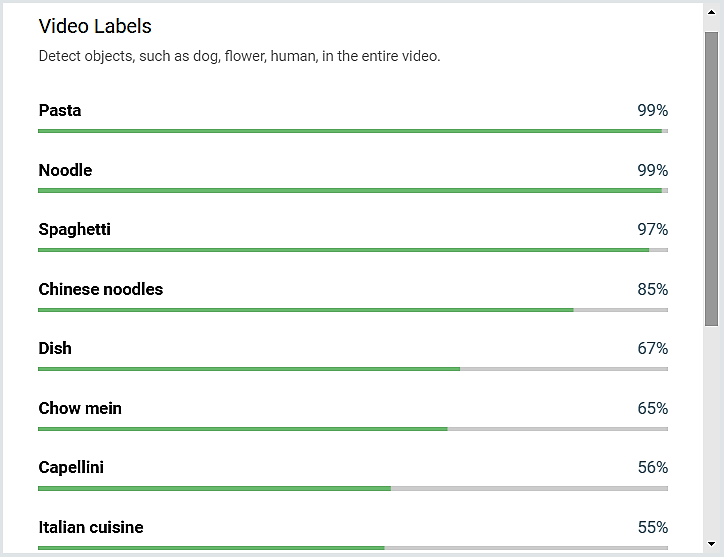}\vspace{0cm}
		\caption{Video labels of the manipulated video, where an image of a food plate is inserted once every two seconds.}
	\end{subfigure}\hspace{1cm}
	\begin{subfigure}{.4\textwidth}
		\centering
		\includegraphics[width=1\linewidth]{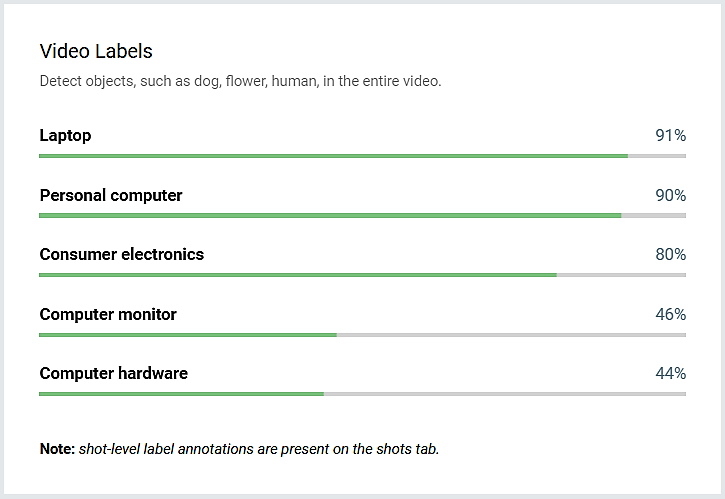}\vspace{0.55cm}
		\caption{Video labels of the manipulated video, where an image of a laptop is inserted once every two seconds.}
	\end{subfigure}
	\caption{The results of the image insertion attack for changing the video labels of the sample video ``Animals.mp4,'' provided by the demonstration website of the Google's Cloud Video Intelligence~\cite{cloud}.}
	\label{fig:animals}
\end{figure*}

\begin{figure*}[h]
	\centering
	\begin{subfigure}{.8\textwidth}
		\centering
		\includegraphics[width=1\linewidth]{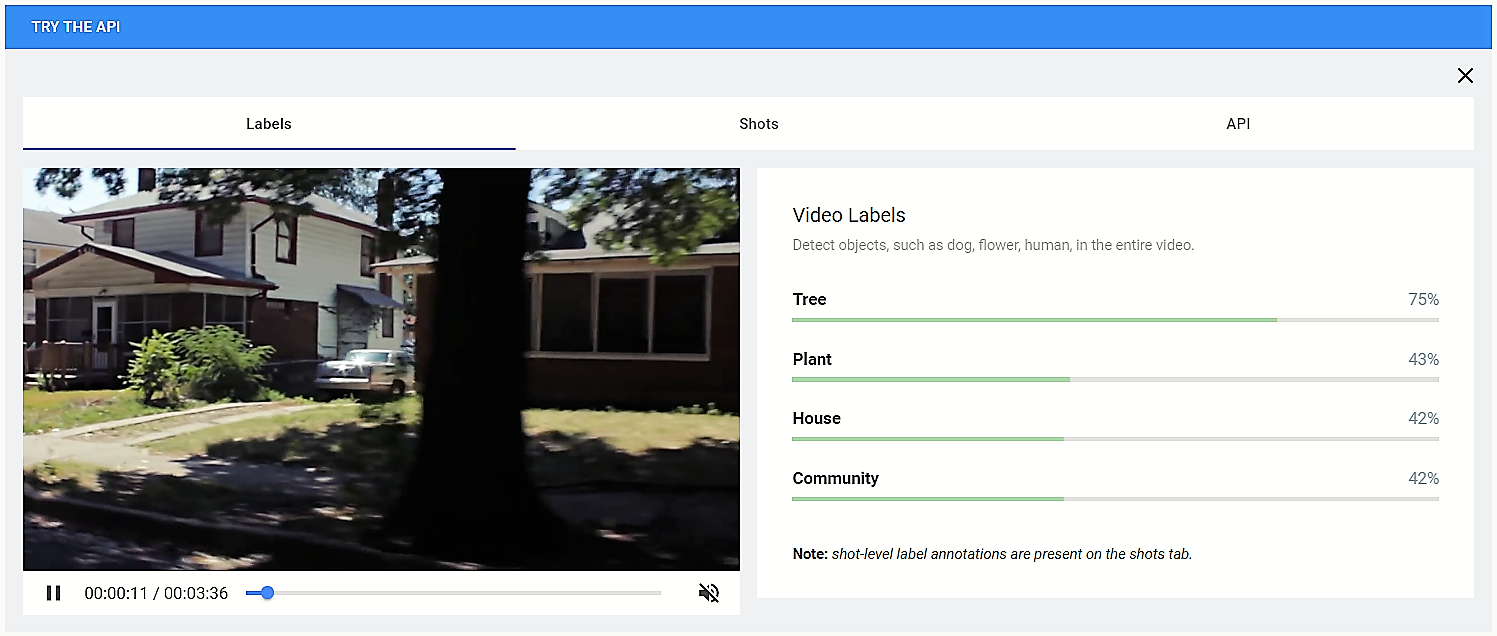}
		\caption{Video labels generated by API for the original video.}
	\end{subfigure}\\\vspace{0.5cm}
	\begin{subfigure}{.4\textwidth}
		\centering
		\includegraphics[width=1\linewidth]{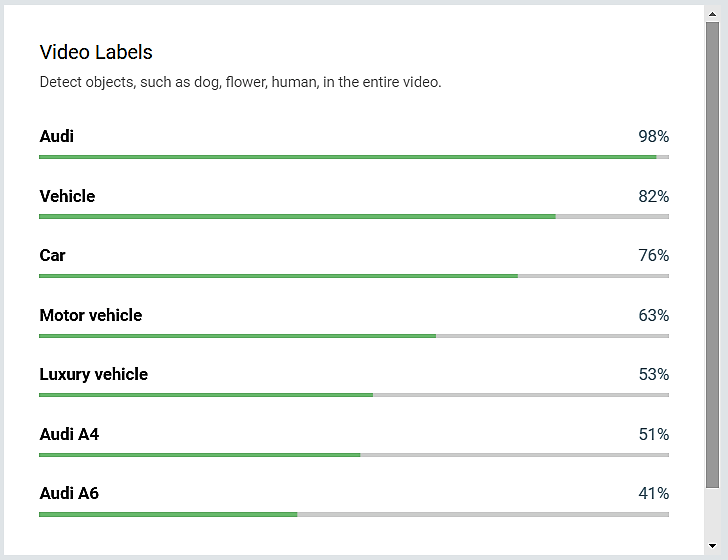}\vspace{0.0cm}
		\caption{Video labels of the manipulated video, where an image of a car is inserted once every two seconds.}
	\end{subfigure}\hspace{1cm}
	\begin{subfigure}{.4\textwidth}
		\centering
		\includegraphics[width=1\linewidth]{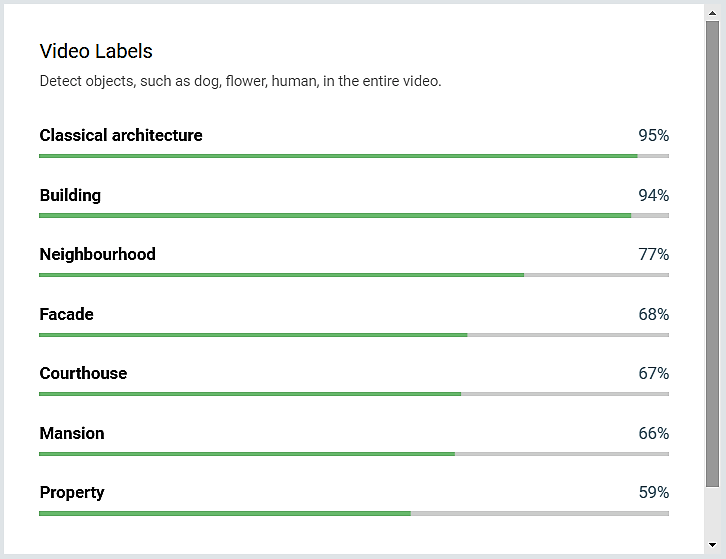}\vspace{0.0cm}
		\caption{Video labels of the manipulated video, where an image of a building is inserted once every two seconds.}
	\end{subfigure}\\\vspace{0.5cm}
	\begin{subfigure}{.4\textwidth}
		\centering
		\includegraphics[width=1\linewidth]{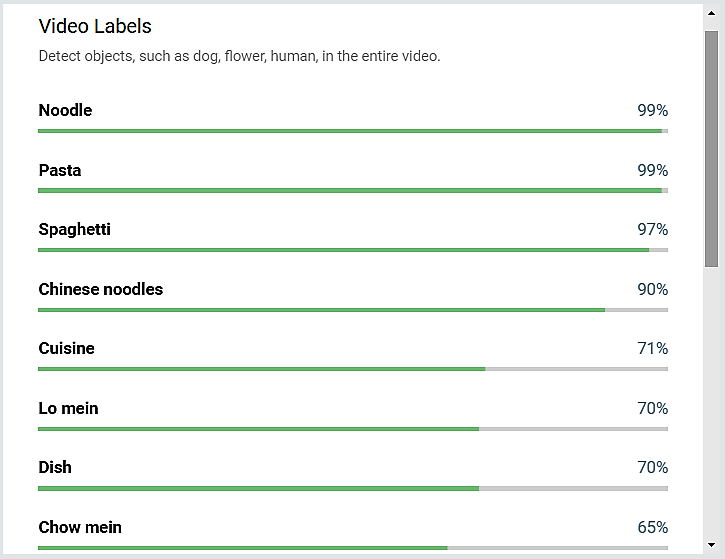}\vspace{0.0cm}
		\caption{Video labels of the manipulated video, where an image of a food plate is inserted once every two seconds.}
	\end{subfigure}\hspace{1cm}
	\begin{subfigure}{.4\textwidth}
		\centering
		\includegraphics[width=1\linewidth]{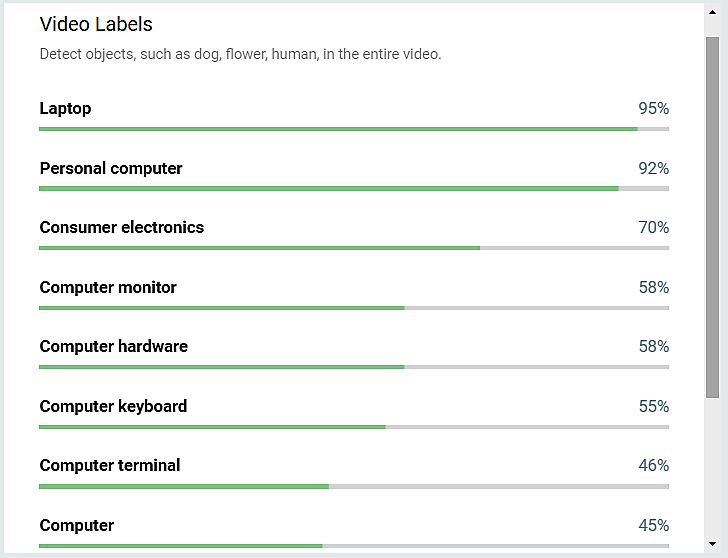}\vspace{0.05cm}
		\caption{Video labels of the manipulated video, where an image of a laptop is inserted once every two seconds.}
	\end{subfigure}
	\caption{The results of the image insertion attack for changing the video labels of the sample video ``GoogleFiber.mp4,'' provided by the demonstration website of the Google's Cloud Video Intelligence~\cite{cloud}.}
	\label{fig:Fiber}
\end{figure*}

\begin{figure*}[h]
	\centering
	\begin{subfigure}{.8\textwidth}
		\centering
		\includegraphics[width=1\linewidth]{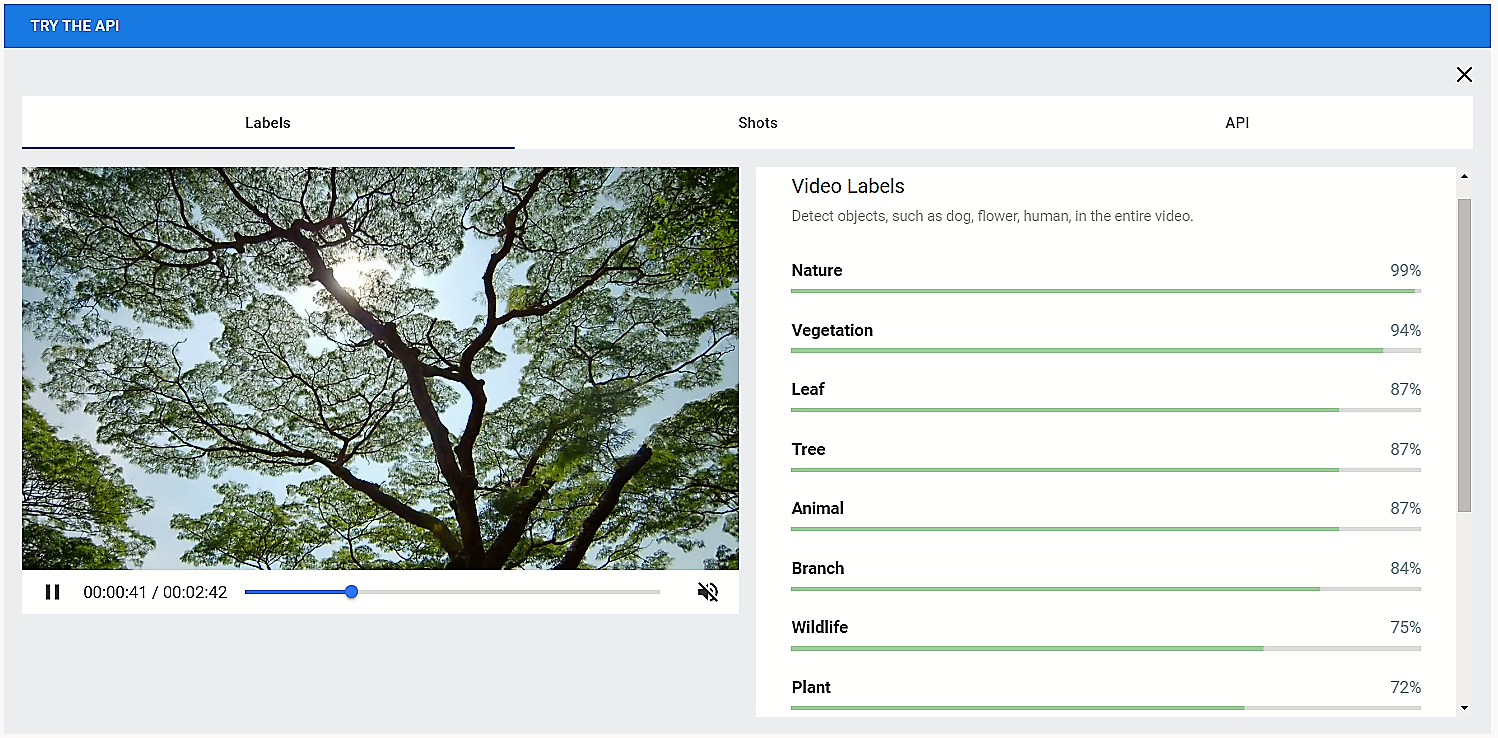}
		\caption{Video labels generated by API for the original video.}
	\end{subfigure}\\\vspace{0.5cm}
	\begin{subfigure}{.4\textwidth}
		\centering
		\includegraphics[width=1\linewidth]{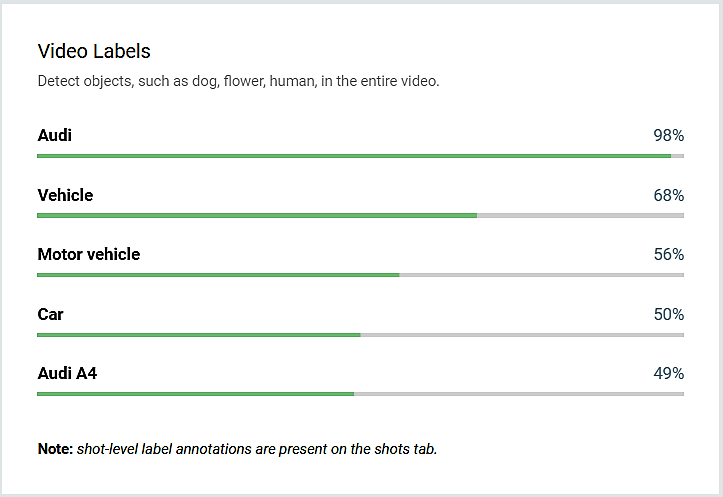}\vspace{0.55cm}
		\caption{Video labels of the manipulated video, where an image of a car is inserted once every two seconds.}
	\end{subfigure}\hspace{1cm}
	\begin{subfigure}{.4\textwidth}
		\centering
		\includegraphics[width=1\linewidth]{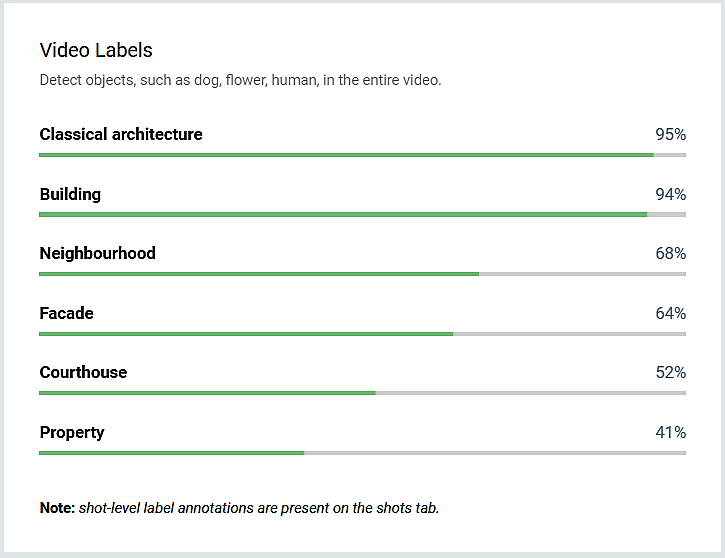}\vspace{0cm}
		\caption{Video labels of the manipulated video, where an image of a building is inserted once every two seconds.}
	\end{subfigure}\\\vspace{0.5cm}
	\begin{subfigure}{.4\textwidth}
		\centering
		\includegraphics[width=1\linewidth]{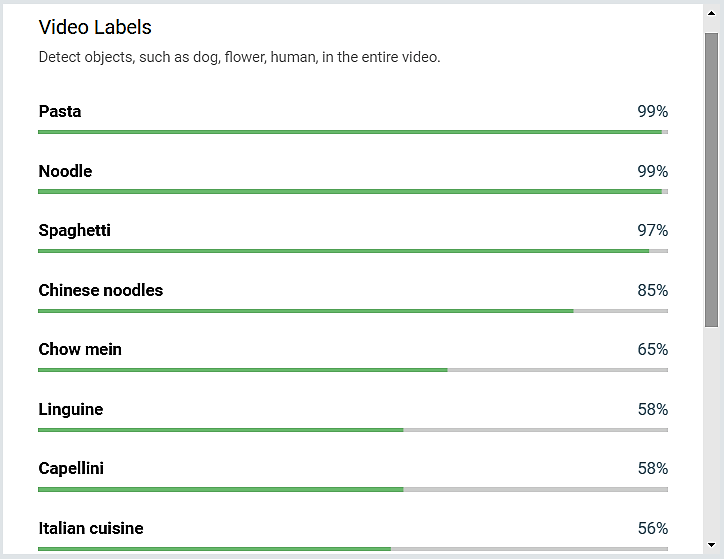}\vspace{0cm}
		\caption{Video labels of the manipulated video, where an image of a food plate is inserted once every two seconds.}
	\end{subfigure}\hspace{1cm}
	\begin{subfigure}{.4\textwidth}
		\centering
		\includegraphics[width=1\linewidth]{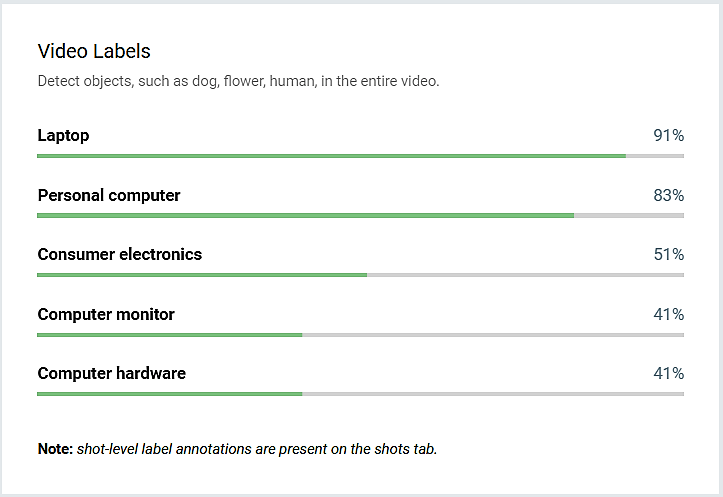}\vspace{0cm}\vspace{0.6cm}
		\caption{Video labels of the manipulated video, where an image of a laptop is inserted once every two seconds.}
	\end{subfigure}
	\caption{The results of the image insertion attack for changing the video labels of the sample video ``JaneGoodall.mp4,'' provided by the demonstration website of the Google's Cloud Video Intelligence~\cite{cloud}.}
	\label{fig:JaneGoodall}
\end{figure*}

Table~\ref{table} provides the API's output for the video labels (the table shows only the label with the highest confidence score). As can be seen, regardless of the video content, the API returns a video label, with a very high confidence score, that exactly matches the corresponding inserted images. 
Figure~\ref{fig:animals} shows the results in more details, providing the screenshots of the video annotations for the sample video ``Animals.mp4'' and the four versions, each manipulated with one of the images presented in Figure~\ref{fig:images}. 
The results show that, while the API can accurately annotate the original video, for the manipulated videos it {\it only} outputs the labels which are related to the inserted image. Figures~\ref{fig:Fiber} and~\ref{fig:JaneGoodall} show similar experiments with the ``GoogleFiber.mp4'' and ``JaneGoodall.mp4'' videos, respectively.

\begin{figure*}[h]
	\centering
	\begin{subfigure}{.775\textwidth}
		\centering
		\includegraphics[width=1\linewidth]{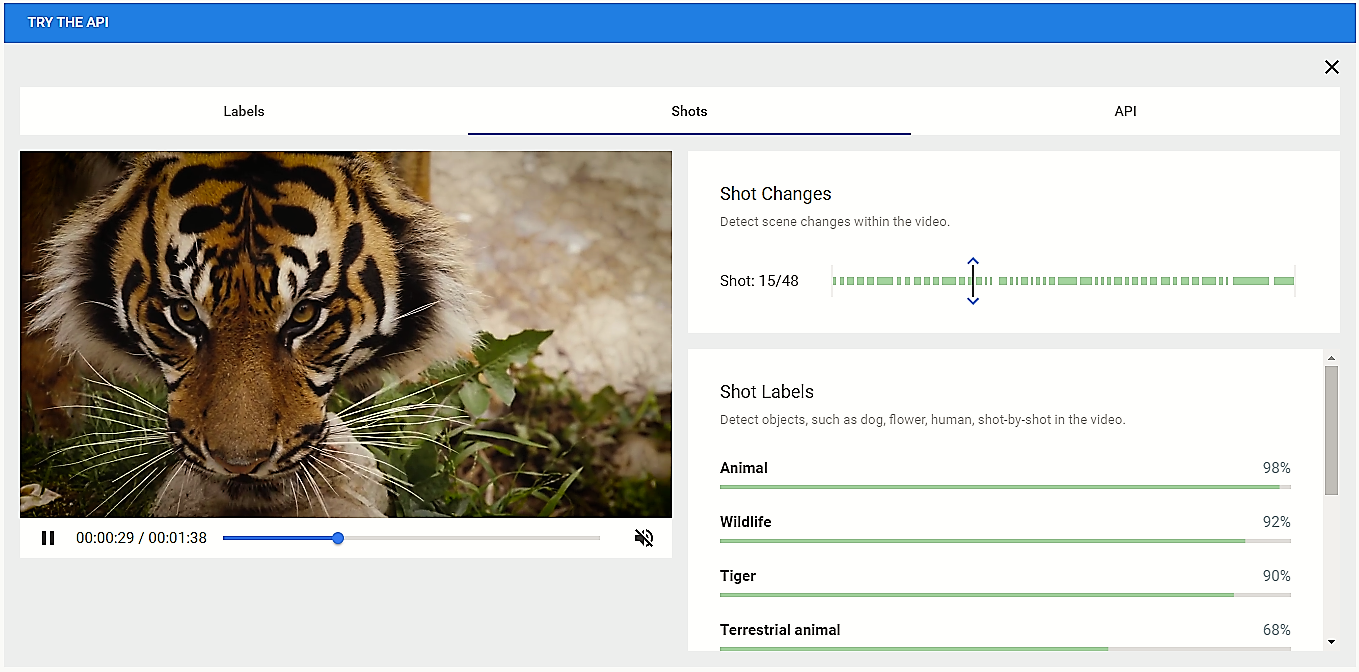}
		\caption{Shot labels generated by API for the original video.}
	\end{subfigure}\\\vspace{0.5cm}
	\begin{subfigure}{.385\textwidth}
		\centering
		\includegraphics[width=1\linewidth]{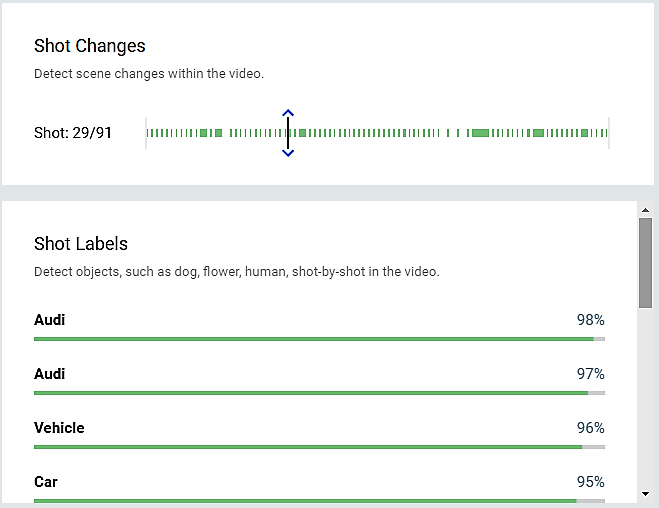}
		\caption{Shot labels of the manipulated video, where an image of a car is inserted once every second.}
	\end{subfigure}\hspace{1cm}
	\begin{subfigure}{.385\textwidth}
		\centering
		\includegraphics[width=1\linewidth]{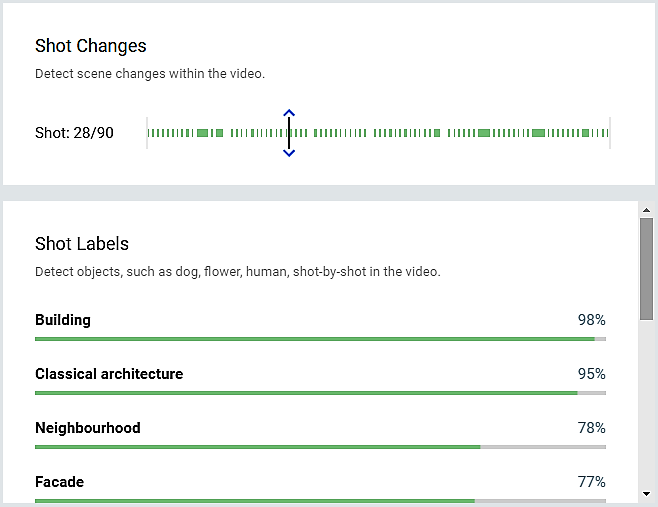}\vspace{0cm}
		\caption{Shot labels of the manipulated video, where an image of a building is inserted once every second.}
	\end{subfigure}\\\vspace{0.5cm}
	\begin{subfigure}{.385\textwidth}
		\centering
		\includegraphics[width=1\linewidth]{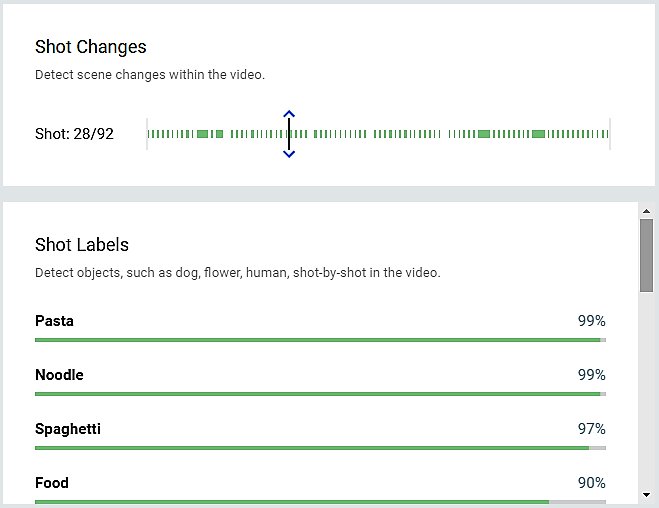}\vspace{0cm}
		\caption{Shot labels of the manipulated video, where an image of a food plate is inserted once every second.}
	\end{subfigure}\hspace{1cm}
	\begin{subfigure}{.385\textwidth}
		\centering
		\includegraphics[width=1\linewidth]{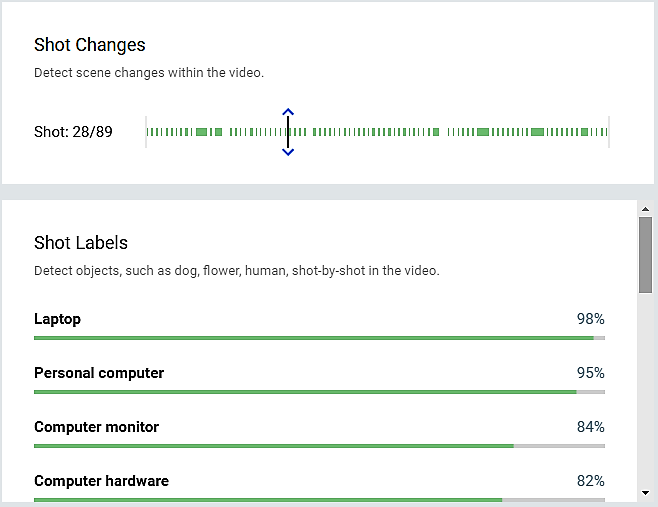}\vspace{0cm}
		\caption{Shot labels of the manipulated video, where an image of a laptop is inserted once every seconds.}
	\end{subfigure}
	\caption{The results of the image insertion attack for changing the shot labels of the sample video ``Animals.mp4,'' provided by the demonstration website of the Google's Cloud Video Intelligence~\cite{cloud}. While the figures shows the results only for one shot, we verified that the attack succeeds to change {\it all} the shot labels to the labels of inserted image. Note that the periodic image insertion also completely alters the {\it shot changes} of the video, returned by the API.}
	\label{fig:shot}
\end{figure*}

We performed similar experiments for changing the video shot labels returned by the API. Note that shot labels provide a detailed description of the individual scenes within the video; therefore, compared to changing the video labels, it is more challenging to change all the shot labels, while maintaining a low image insertion rate. However, we found that by inserting one image per second, resulting in an image insertion rate of $0.04$ for the frame rate of $25$, {\it all} the shot labels returned by the API are related to the inserted image. Figures~\ref{fig:shot} shows the screenshots of the shot labels for the original video ``Animals.mp4'' and the four manipulated versions, each with one of the inserted images. While the figure shows the results only for one shot, we verified that the attack succeeds to change all the shot labels to the labels of inserted image. 
Moreover, it can be seen that the proposed image insertion attack completely alters the pattern of the {\it shot changes} of the video, returned by the API.

\section{Discussion}

Many applications can benefit from automated video search and summarization. For example, in video surveillance, one needs to search many hours of videos for a specific event. Also, some Internet platforms, such as YouTube and Facebook, require to process enormous amounts of video files every day, for video recommendation and to block the videos with illegal contents. 
The Google's Cloud Video Intelligence API is designed to enable the developers to quickly search the video contents, just as text documents. Hence, it has the potential to transform the video analysis field to the point that users can search for a particular event and get related videos along with the exact timings of the events within the videos.

However, we showed that the API has certain security weaknesses. Specifically, an adversary can insert an image, periodically and at a very low rate, into the video in a way that all the generated shot labels are about the inserted image. Such vulnerability seriously undermines the applicability of the API in adversarial environments. For example, one can upload a manipulated video which contains adversarial images related to a specific event, and the API wrongly suggests it to users who asked for videos from the event. Furthermore, an adversary can bypass a video filtering system by inserting a benign image into a video with illegal contents.

Note that we could deceive the Google's Cloud Video Intelligence API, without having any knowledge about the learning algorithms, video annotation algorithms or the cloud computing architecture used by the API. 
That is, we developed an approach for deceiving the API, by only querying the system with different inputs. Through experiments, we showed that the attack is consistently successful with different videos and images. 
The success of the image insertion attack shows the importance of designing the system to work equally well in adversarial environments.

\section{Conclusion}

In this paper, we showed that the Google's Cloud Video Intelligence API can be easily deceived by an adversary without compromising the system or having any knowledge about the specific details of the algorithms used. In essence, we found that an adversary can slightly manipulate a video by inserting an image periodically into it, such that the API returns only the labels that are related to the inserted image. 

\vspace{0.35cm}
\noindent{\bf \large Acknowledgments}

\vspace{0.1cm}
\noindent This work was supported by ONR grants N00014-14-1-0029 and N00014-16-1-2710, ARO grant W911NF-16-1-0485 and NSF grant CNS-1446866.

\vspace{0.1cm}
\noindent The views presented in this paper are those of the authors and do not reflect the position of sponsoring agencies.

{\small
\bibliographystyle{ieeetr}
\bibliography{Main}
}

\end{document}